# Propagation using Chain Event Graphs

**Peter A. Thwaites**
Statistics Dept.
University of Warwick
Coventry UK CV4 7AL

**Jim Q. Smith**
Statistics Dept
University of Warwick
Coventry UK CV4 7AL

**Robert G. Cowell**
Cass Business School
City University
London EC1Y 8TZ

## Abstract

A Chain Event Graph (CEG) is a graphical model which is designed to embody conditional independencies in problems whose state spaces are highly asymmetric and do not admit a natural product structure. In this paper we present a probability propagation algorithm which uses the topology of the CEG to build a *transporter* CEG. Intriguingly, the transporter CEG is directly analogous to the triangulated Bayesian Network (BN) in the more conventional junction tree propagation algorithms used with BNs. The propagation method uses factorization formulae also analogous to (but different from) the ones using potentials on cliques and separators of the BN. It appears that the methods will be typically more efficient than the BN algorithms when applied to contexts where there is significant asymmetry present.

## 1 INTRODUCTION

Based on an event tree, a Chain Event Graph (CEG) is a more expressive alternative to a discrete Bayesian Network (BN), embodying collections of conditional independence statements in its topology. In Anderson and Smith (2008) it is shown not only how asymmetries in a problem's sample space can be represented explicitly through the topology of its CEG, but also how it can express a much wider range of types of conditional independence statement not simultaneously expressible through a single BN. As with the BN, the CEG of an hypothesised model can be interrogated using natural language before the graph is embellished with probabilities. In Thwaites and Smith (2006) and Riccomagno and Smith (2005) we demonstrate how the CEG can also be used to represent and analyse various causal hypotheses. In this paper we continue the development of CEGs by demonstrating how the graph provides a useful structure for fast probability propagation in asymmetric models.

It has been noted that the CEG is an especially powerful framework for inference when a probability model is highly asymmetric and elicited through a description of how situations unfold. Although theoretically a BN can be used in this context, the clique probability tables are then very sparse and contain many zeros or repeated probabilities. This impedes fast propagation algorithms and has led to the development of many context specific variants of BNs (Boutilier et al 1996, McAllester et al 2004, Poole and Zhang 2003, Salmeron et al 2000), often based on trees within cliques. These developments provoke the question as to whether a single tree might be used for propagation instead of the BN. Now obviously the event tree itself expresses no conditional independencies in its topology and these independencies are the building blocks of current propagation algorithms. However, unlike the event tree, the CEG expresses a fairly comprehensive collection of conditional independencies. In this paper we demonstrate the surprising fact that there is a direct analogue between a distribution on a BN expressed as a product of potentials supported by a graph of cliques and separators, and propagation algorithms on CEGs using the distributions on the children of the CEG's non-leaf nodes and marginal likelihoods on the vertices themselves. This enables us to develop fast propagation algorithms that use a single graph, the *transporter* CEG – analogous to a triangulated BN – as its framework. This framework is highly efficient for asymmetric/non-product-space contexts, and in particular does not involve propagating zeros in sparse but large probability tables, nor continually repeating the same calculations, which would be the case if we were to use the BN as a framework in this sort of environment with a naive BN propagation algorithm.

In the next section we formally define the transporter CEG $C(T)$ of a hypothesised probability tree $T$. In section 3 we present an algorithm analogous to that of Cowell and Dawid (1992) for a BN where conditional probability tables associated with the children of a given vertex of the CEG take the role of cliques, and vertex probabilies take the role of separators. In section 4 we demonstrate the efficiency of this algorithm with a simple example.

## 2 PROBABILITY TREES AND CHAIN EVENT GRAPHS

Probability trees (Shafer 1996), and their control analogues decision trees, have been found to be a very natural and expressive framework for probability and decision problems, and they provide an excellent framework for describing sample space asymmetry and inhomogeneity in a given context (see for example French and Insua (2000)). We start with an event tree $T$ with vertex set $V(T)$ and (directed) edge set $E(T)$. Henceforth call the tree's non-leaf vertices $\{v\}$ *situations*, and denote this set of vertices $S(T) \subset V(T)$. We can convert an event tree into a probability tree by specifying a transition matrix from its vertices $V(T)$, where the absorbing states correspond to the leaf vertices. Transition probabilities from a situation are zero except for transitions to one of that situation's children. This makes the transition matrix upper triangular. Such a matrix would look like the one in Table 1 which shows part of the matrix for the problem described in Example 1. Note that each transition probability can be identified by an edge on the tree.

Table 1: Part of the transition matrix for Example 1

| | $v_0$ | $v_1$ | $v_2$ | $v_3$ | $v_4^1$ | $v_4^2$ | $v_4^3$ | $v_5^1$ | $v_5^2$ | $\cdots$ | $v_\infty^1$ | $\cdots$ |
|---|---|---|---|---|---|---|---|---|---|---|---|---|
| $v_0$ | 0 | $\theta_1$ | $\theta_2$ | $\theta_3$ | 0 | 0 | 0 | 0 | 0 | $\cdots$ | 0 | $\cdots$ |
| $v_1$ | 0 | 0 | 0 | 0 | $\theta_5$ | 0 | 0 | 0 | 0 | $\cdots$ | $\theta_4$ | $\cdots$ |
| $v_2$ | 0 | 0 | 0 | 0 | 0 | $\theta_6$ | 0 | $\theta_7$ | 0 | $\cdots$ | 0 | $\cdots$ |
| $v_3$ | 0 | 0 | 0 | 0 | 0 | 0 | $\theta_8$ | 0 | $\theta_9$ | $\cdots$ | 0 | $\cdots$ |
| $\vdots$ | | | | | | | | | $\vdots$ | | $\vdots$ | |

One way of seeing conditional independence statements on a BN is as identities in certain vectors of conditional probabilties – explicitly those probability vectors associated with different ancestor configurations but the same parent configuration of a variable in the BN (Riccomagno and Smith 2007). There is a large class of models where the probabilities in some of the rows of the transition matrix can be identitifed with each other. The CEG is a topological representation of this class of models, and the *transporter* CEG defined below is a subgraph of the CEG.

Let $T(v_i)$, $i = 1, 2$ be the unique subtrees whose roots are the situations $v_i$, and which contain all vertices after $v_i$ in $T$. Say $v_1$ and $v_2$ are in the same *position* $w$ if:

1. the trees $T(v_1)$ and $T(v_2)$ are topologically identical.
2. there is a map between $T(v_1)$ and $T(v_2)$ such that the edges in $T(v_2)$ are annotated, under that map, by the same (possibly unknown) probabilities as the corresponding edges in $T(v_1)$.

It is easily checked that the set $W(T)$ of positions $w$ partitions $S(T)$. Furthermore, somewhat more subtlely, if $v_1, v_2 \in w$ and $v_{ij} \in V(T(v_i))$, then the vertex sets of $T(v_i)$ $i = 1, 2$ are mapped on to each other by this map, and $v_{ij} \in w_j$ $i = 1, 2$ for some position $w_j$ (providing $v_{ij}$ is not a leaf-vertex in either subtree). For details of this property see Anderson and Smith (2008).

We now draw a new graph to depict both the sample space of $T$ and certain conditional independence statements. The *transporter* CEG $C(T)$ is a directed graph whose vertices $W(C(T))$ are $W(T) \cup \{w_\infty\}$. There is an edge ($e \in E(C(T))$) from $w_1$ to $w_2 \neq w_\infty$ for each situation $v_2 \in w_2$ which is a child of a fixed representative $v_1 \in w_1$ for some $v_1 \in S(T)$, and an edge from $w_1$ to $w_\infty$ for each leaf node $v \in V(T)$ which is a child of some fixed representative $v_1 \in w_1$ for some $v_1 \in S(T)$. The transporter CEG (henceforth labelled simply as $C$) is the subgraph of a CEG (defined in Anderson and Smith (2008)) where all undirected edges in the CEG are omitted. The relationship between the transporter CEG and the CEG is directly analogous to the relationship between a triangulated BN and the original BN. Certain conditional independence statements that can be lost through conditioning are simply forgotten so that an homogeneous propagation algorithm can be constructed on the basis of the enduring conditional independencies. Unlike the BN, this CEG can have many edges between two vertices and always has a single sink vertex $w_\infty$. Although typically having many fewer vertices than $T$, it retains a depiction of the sample space structure of $T$. Thus it is easy to check that the set of root to leaf paths of the tree (representing the set of all possible unfoldings of the history of a unit) are in one to one correspondence with the set of root to sink paths on the transporter CEG. The CEG-construction process is illustrated in Example 1.

**Example 1**

Consider the tree in Figure 1, which has 16 atoms (root-to-leaf paths). Note that as the subtrees rooted in the vertices $\{v_4^i\}$ are the same, and those rooted in $\{v_5^i\}$ are the same, the distribution on the tree can be stored using 7 conditional tables which contain 16 (9 free) probabilities.

Our transporter CEG (Figure 2) is produced by combining the vertices $\{v_4^i\}$ into one *position* $w_4$, the ver-tices $\{v_5^i\}$ into one position $w_5$, the vertices $\{v_6^i\}$ into one position $w_6$, and all leaf-vertices into a single sink-node $w_\infty$. The full CEG for our example is *simple* – it has no undirected edges, and is identical to the transporter CEG $C$. For a *simple* CEG, **all** the conditional independencies inherent in the problem are conveyed by the transporter CEG.

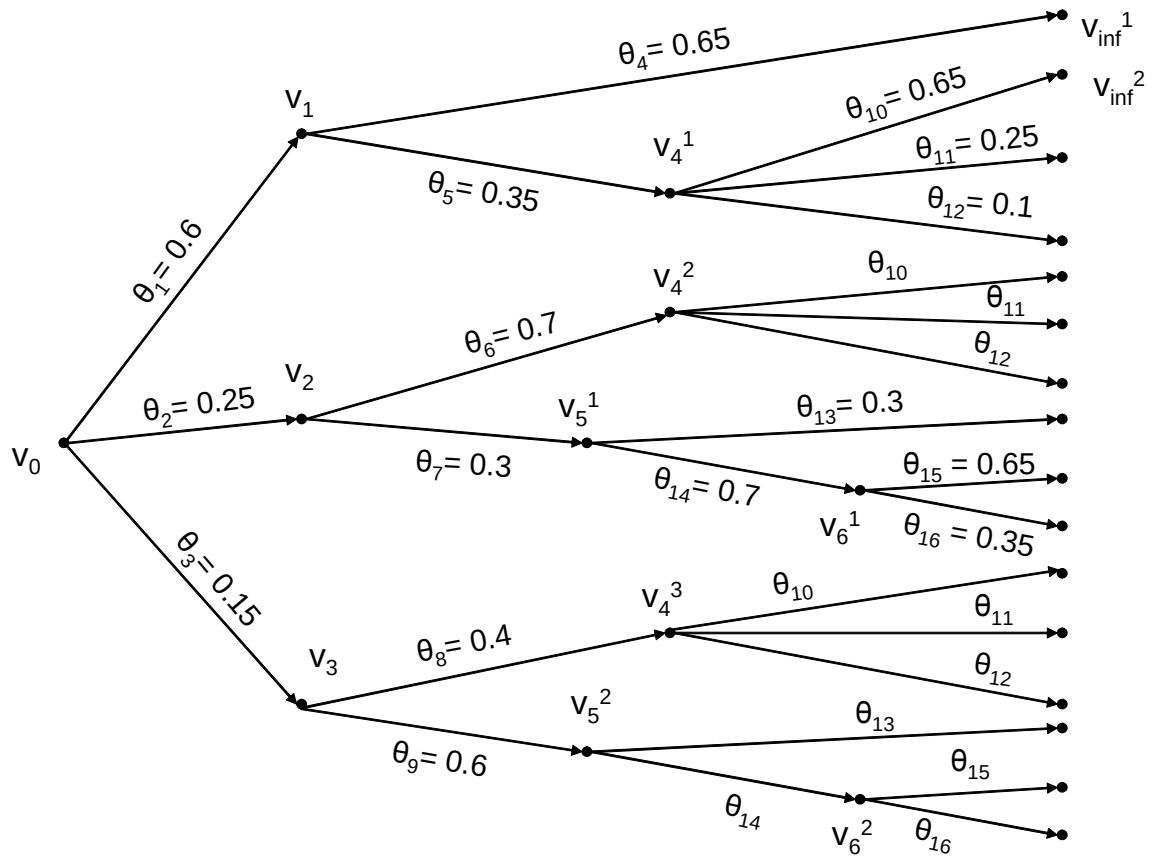

Figure 1: Tree for Example 1

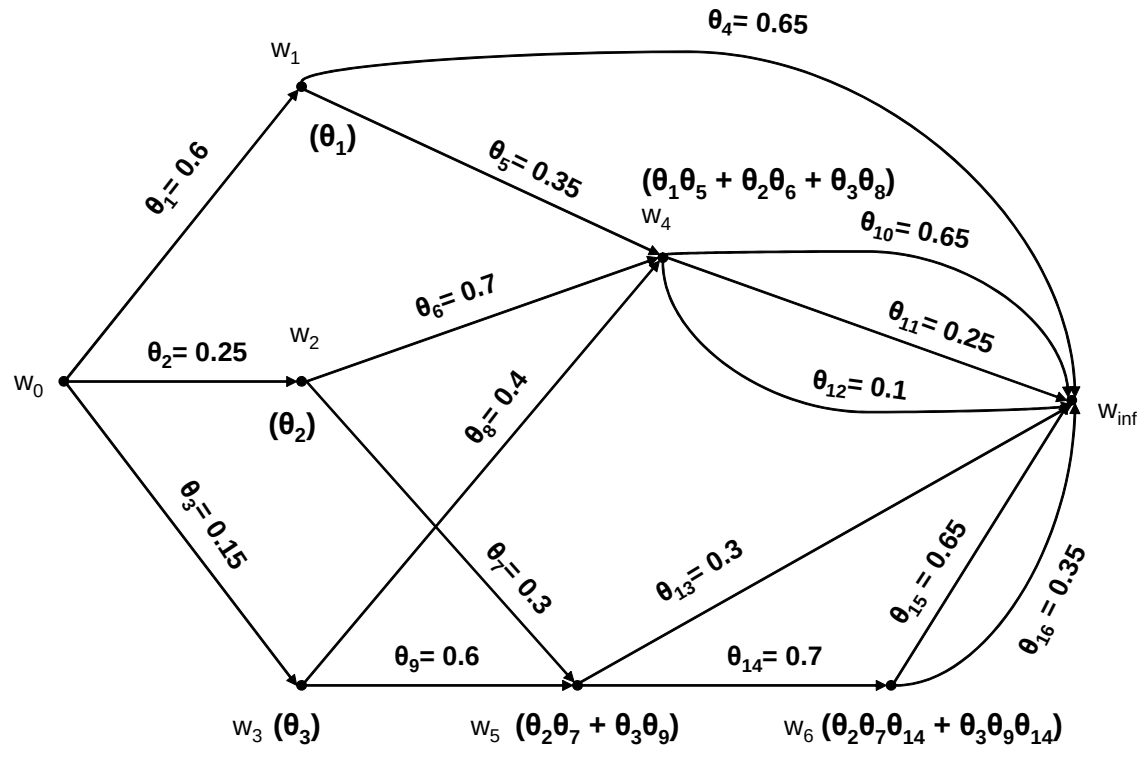

Figure 2: Transporter CEG for Example 1

Figure 2 shows the probabilities of reaching each position $w$ (the event *reaching* $w$, denoted $\Lambda(w)$, is the union of all root-to-sink paths passing through $w$). It also shows each edge-probability $\pi_e(w' \mid w)$ $(= \pi(\Lambda(e(w,w')) \mid \Lambda(w))$, where $\Lambda(e(w,w'))$ is the union of all root-to-sink paths utilising the edge $e(w,w')$ ).

The problem represented by the tree in Figure 1 is asymmetric in that not all the root-to-leaf paths are of the same length, and also in the local structure associated with its vertices. We do not know whether the vertices $\{v_4^i\}$ are related in any contextual way to the vertices $\{v_5^i\}$ or $\{v_6^i\}$, and hence we cannot obviously define variables on the sigma-algebra of the tree to allow us to represent the problem as a BN. Even supposing we were able to represent the problem in such a way, the conditional independencies embodied in the problem (and in our transporter CEG) cannot be efficiently coded in a BN without introducing tables with many zeros. Consequently, even in this very simple example we have efficiency gains in storing this distribution over using a saturated model, a BN, or a tree.

# 3 A SIMPLE PROPAGATION ALGORITHM

## 3.1 THE FRAMEWORK

To specify the joint distribution of all random variables measurable with respect to a CEG we simply need to specify the vector of conditional probability mass functions associated with each of its positions. The first step of our propagation algorithm is analogous to the triangulation step for a BN, which allows us to retain all conditional independence properties at the cost of a possible loss of efficiency. To do this we ignore conditional independence statements coded by the undirected edges of the CEG and work only with the subgraph consisting of its positions, together with its directed edges, but not its undirected edges – our transporter CEG $C$.

For each position $w \in W = W(C)\backslash\{w_\infty\}$ we store a vector of probabilites $\boldsymbol{\pi}(w) = \{\pi_e(w' \mid w) \mid e(w,w') \in E(w)\}$ where $E(w) \subset E(C)$ is the set of all edges emanating from $w$. $\boldsymbol{\pi}(w)$ is of course a conditional probability distribution. We let $X(w)$ be the random variable taking values on $\{1, 2, \ldots, n(E(w))\}$ (where $n(E(w))$ is the number of edges emanating from $w$) whose probability mass function is given by the components of $\boldsymbol{\pi}(w)$ taken in order. The positions $w \in W$ take the role of the cliques in a triangulated BN, whilst the vectors $\{\boldsymbol{\pi}(w) \mid w \in W\}$ are analogous to the clique probability tables.

We can now specify the probability $\pi_\lambda$ of every atom $\lambda$ (a root to sink path of $C$, of length $n(\lambda)$ ) as a function of $\{\boldsymbol{\pi}(w) \mid w \in W\}$ and $C$. If:

$$\lambda = (w_0 = w_\lambda[0], e_\lambda[1], w_\lambda[1], \ldots, e_\lambda[n(\lambda)], w_\infty)$$

then

$$\pi_\lambda = \prod_{i=1}^{n(\lambda)} \pi(e_\lambda[i])$$

where $\pi(e_\lambda[i])$ is a component of the probability vector $\boldsymbol{\pi}(w_\lambda[i-1])$, $1 \le i \le n(\lambda)$. It follows that the distribution of any random variable measurable with respect to $C$ can be calculated from $\{\boldsymbol{\pi}(w) \mid w \in W\}$.

## 3.2 COMPATIBLE OBSERVATIONS

Recall that propagation algorithms for BNs based on triangulation are only designed to propagate information that can be expressed in the form $O(\mathbf{A}) = \{X_j \in A_j\}$ for some subsets $\{A_j\}$ of the sample spaces of $\{X_j\}$ the vertex-variables of the BN. Propagating information about the value of some general function of the vertex variables using local message passing is not generally possible, because conditioning on the values of such a function can destroy the conditional independencies on which the local steps of the propagation algorithm depend for their validity.

In the same way the types of observation we can efficiently propagate using $C$ and $\{\boldsymbol{\pi}(w) \mid w \in W\}$ needs to be constrained. In general an observation can be identified with a subset $\Lambda$ of the set of all root to sink paths $\{\lambda\}$. The most obvious constraining assumption on $\Lambda$ (and the one we will henceforth make in this paper) about what we might learn is that our observation $\Lambda$ can be identified with having learned that $\{X(w) \in A(w)\}$ for some subsets $\{A(w)\}$ of the sample spaces of the position random variables $\{X(w)\}$. Call such a set $C-$*compatible*. Note that $\Lambda$ is $C-$compatible if and only if there exists possibly empty subsets $\{E_\Lambda(w) \mid w \in W\}$ such that

$$\Lambda = \{\lambda \mid e_\lambda \in E_\Lambda(w) \text{ for some } w \in W, \text{ for each edge } e_\lambda \text{ on the path } \lambda \text{ in } C\}$$

So we can identify a compatible observation with the set of edges $E_\Lambda = \bigcup_{w \in W} E_\Lambda(w) \subset E(C)$. We note that the set of compatible observations is large and in particular when the CEG is expressible as a BN contains all sets of the form $O(\mathbf{A})$ defined above.

**Example 2**

Consider:

$$\Lambda = \{\lambda \mid e_\lambda \in \{e_1(w_0, w_1), e_2(w_0, w_2), e_4(w_1, w_\infty), e_5(w_1, w_4), e_6(w_2, w_4), e_7(w_2, w_5), e_{10}(w_4, w_\infty), e_{11}(w_4, w_\infty), e_{14}(w_5, w_6), e_{15}(w_6, w_\infty)\}\}$$

This corresponds to all the root-to-sink paths in the subgraph of $C$ given in Figure 3.

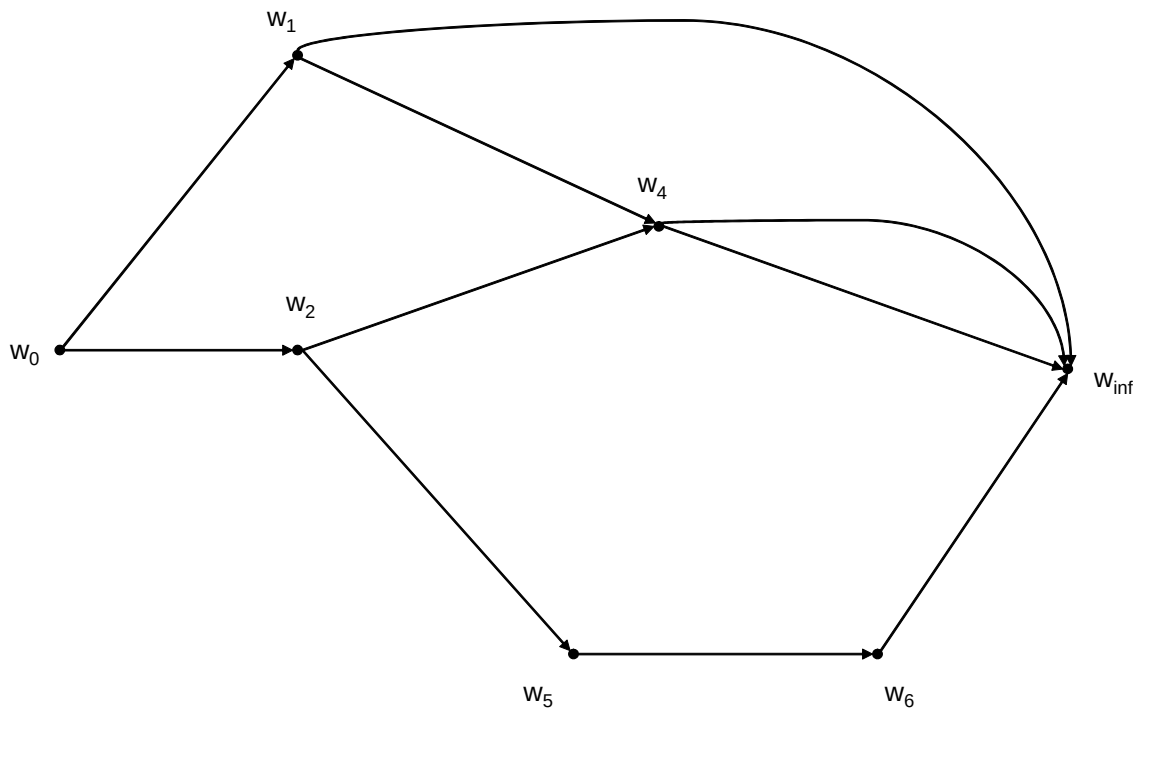

Figure 3: Subgraph for event $\Lambda$ in Example 2

### 3.3 MESSAGE PASSING FROM COMPATIBLE OBSERVATIONS ON A CEG

The message passing algorithm is a function from the original probabilities $\{\boldsymbol{\pi}(w) \mid w \in W\}$ to revised probabilities on the same graph $\{\hat{\boldsymbol{\pi}}(w) \mid w \in W\}$ conditional on the observation $\Lambda$. Note that once edge-probabilities have been revised, the resulting graph may not be a *minimal* CEG (in that we may have vertices within the graph which are the roots of identical sub-graphs). It is possible (although unnecessary for information-propagating purposes) to add a further algorithm step to produce a minimal CEG if this is required. This step ensures that any vertices that are equivalent are combined into a single position.

Messages are passed from the terminal edges backwards through the transporter CEG along neighbouring edges until reaching the root in a *collect* step giving a new pair $\{\boldsymbol{\tau}(w), \Phi(w) \mid w \in W\}$. We then move forward from the root producing revised $\{\hat{\boldsymbol{\pi}}(w) \mid w \in W\}$. Let $W(-1)$ denote the set of positions all of whose outgoing edges terminate in $w_\infty$ in $C$.

1. For any edge $e(w, w_\infty)$ such that $w \in W(-1)$, set the *potential* $\tau_e(w_\infty \mid w) = 0$ if $e(w, w_\infty) \notin E_\Lambda$, and $\tau_e(w_\infty \mid w) = \pi_e(w_\infty \mid w)$ if $e(w, w_\infty) \in E_\Lambda$. Let the *emphasis*:

$$\Phi(w) = \sum_{e \in E(w)} \tau_e(w_\infty \mid w)$$

   Say that $w_\infty$ and each of these positions is *accommodated.*

2. For any position $w$ all of whose children are accommodated, and edge $e(w, w')$, set the *potential* $\tau_e(w' \mid w) = 0$ if $e(w, w') \notin E_\Lambda$, and $\tau_e(w' \mid w) = \pi_e(w' \mid w)\ \Phi(w')$ if $e(w, w') \in E_\Lambda$. Let the *emphasis*:

$$\Phi(w) = \sum_{e \in E(w)} \tau_e(w' \mid w)$$

   Say that $w$ is acommodated.

3. Repeat step 2 until all $w \in W$ are accommodated. This completes the collect steps.

4. For all $w \in W$, set:

$$\hat{\boldsymbol{\pi}}(w) = \mathbf{0} \quad \text{if } \boldsymbol{\tau}(w) = \mathbf{0}$$
$$\hat{\boldsymbol{\pi}}(w) = \frac{\boldsymbol{\tau}(w)}{\Phi(w)} \quad \text{if } \boldsymbol{\tau}(w) \neq \mathbf{0}$$

   where $\boldsymbol{\tau}(w) = \{\tau_e(w' \mid w) \mid e(w, w') \in E(w)\}$.

Clearly we have that:

$$\hat{\pi}_e(w' \mid w) = 0 \quad \text{if } e(w, w') \notin E_\Lambda$$
$$\hat{\pi}_e(w' \mid w) = \frac{\tau_e(w' \mid w)}{\Phi(w)} \quad \text{if } e(w, w') \in E_\Lambda$$

A proof of these results is given in the appendix.

Note that as we move forward through the graph the updated probabilities of $\mu(w_0, w)$ subpaths will be of the form:

$$\hat{\pi}_\mu(w \mid w_0) = \prod_{i=0} \hat{\pi}_e(w_{i+1} \mid w_i)$$

and we get:

$$\hat{\pi}(\Lambda(w)) = \sum_{\mu \in \{\mu(w_0, w)\}} \hat{\pi}_\mu(w \mid w_0)$$

From the definition of accommodation, the order of these operations (like the perfect order used to update a triangulated BN) depends only on the toplogy of $C$, so it can be set up beforehand.

**Example 3**

Steps 1, 2 and 3 give us the graph in Figure 4. Step 4 gives us the CEG in Figure 5 (note that our CEG is again simple, and also minimal without the need for the additional step previously mentioned).

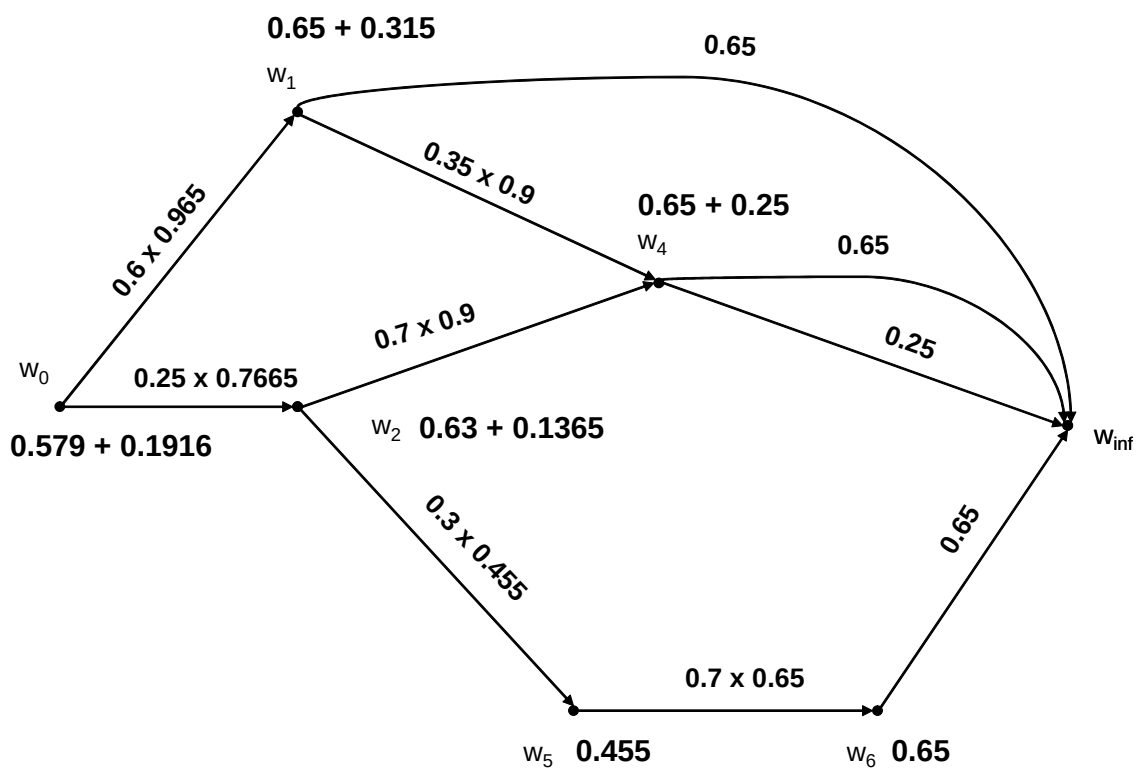

Figure 4: Potentials and emphases added

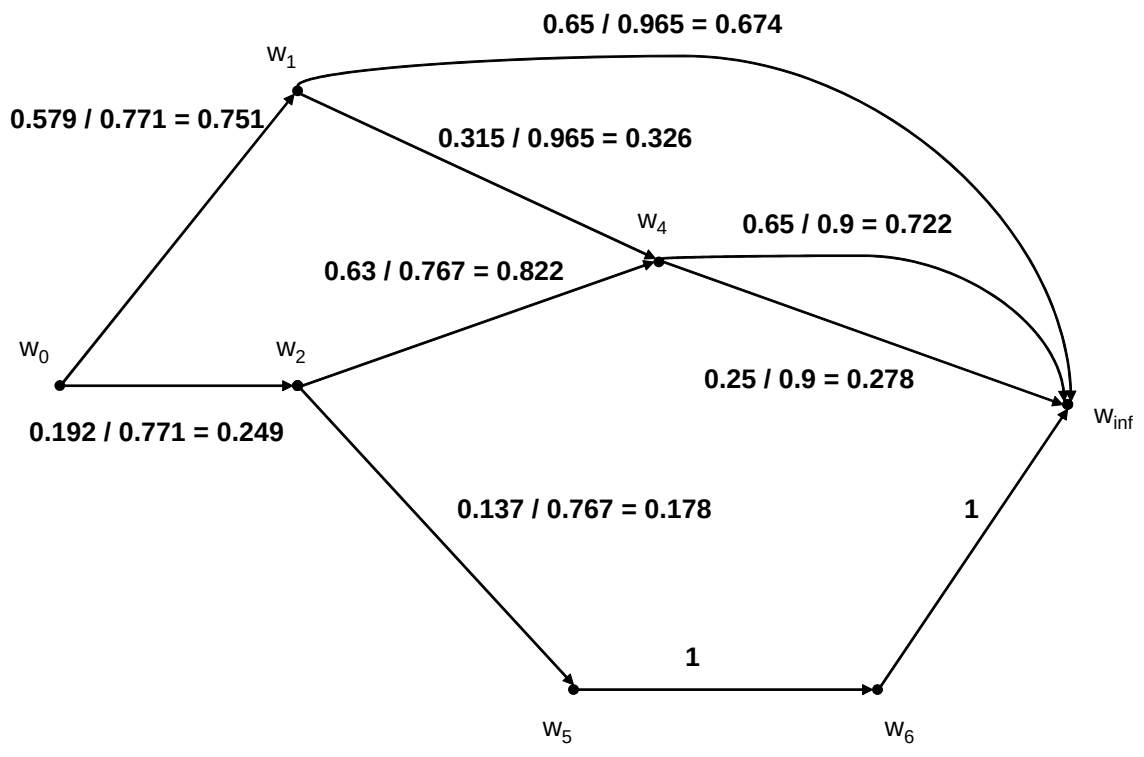

Figure 5: Updated CEG $C_\Lambda$

Note that, in analogy with equation (6) of Cowell and Dawid (1992), the conditional probability of any atom $\lambda = (w_0 = w_\lambda[0], e_\lambda[1], w_\lambda[1], \ldots, e_\lambda[n(\lambda)], w_\infty)$ is given by the invariance formula:

$$\pi(\lambda \mid \Lambda) = \hat{\pi}(\lambda) = \prod_{i=1}^{n(\lambda)} \hat{\pi}(e_\lambda[i]) = \frac{\prod_{i=1}^{n(\lambda)} \tau(e_\lambda[i])}{\prod_{i=0}^{n(\lambda)-1} \Phi(w_\lambda[i])}$$

Also note that at the cost of some computation, we can perform inference on the reduced graph $C_\Lambda$ whose edges $E(C_\Lambda)$ are just the edges $e$ in $E(C)$ with non-zero probabilities $\hat{\pi}(e)$, and whose vertices $W(C_\Lambda)$ are the $w \in W(C)$ for which $\Phi(w) \neq 0$. The non-zero edge and vertex probabilities of $C$ then simply map on to their corresponding edge and vertex probabilities in $C_\Lambda$. Note that, unlike for the BN, any non trivial $C-$compatible observation strictly reduces the number of edges in the edge set after this operation.

A pseudo-code version of our algorithm is provided below:

Let $C(W(C), E(C))$ be a transporter CEG with edges in $E(C)$ having labels $e_i$, $i = 1, 2, \ldots n_e$, such that $i < j \Rightarrow e_i \not\succ e_j$ ($e_i$ does not lie downstream of $e_j$ on any $w_0 \to w_\infty$ path); and positions in $W(C) \backslash \{w_\infty\}$ having labels $w_i$, $i = 0, 1, 2, \ldots m_w$, such that $i < j \Rightarrow w_i \not\succ w_j$. To update the edge-probabilities on $C$ following observation of an event $\Lambda$, do:

(1) Set $A = \phi$
(2) Set $B = \phi$
(3) Set $i = 1$
(4) **Repeat**
  (a) Select $e_i$
  (b) **If** $e_i \in E_\Lambda$, add $e_i$ to $A$
    **otherwise**, set $\hat{\pi}_{e_i} = 0$
  (c) Set $i = i + 1$
 **Until** $i = n_e + 1$
(5) Set $\Phi(w_\infty) = 1$
(6) Set $j = m_w$
(7) **Repeat**
  (a) Select $w_j$
  (b) **Repeat**
    (i) Select $e(w_j, w'_j) \in E(w_j) \cap A$
    (ii) Set $\tau_e(w'_j \mid w_j) = \pi_e(w'_j \mid w_j)\ \Phi(w'_j)$
    (iii) Add $e(w_j, w'_j)$ to $B$
   **Until** $E(w_j) \cap A \subset B$
  (c) Set $\Phi(w_j) = \sum_{e \in E(w_j)} \tau_e(w'_j \mid w_j)$
  (d) Set $j = j - 1$
 **Until** $j = -1$
(8) For each $e(w, w') \in E_\Lambda$, set $\hat{\pi}_e(w' \mid w) = \frac{\tau_e(w' \mid w)}{\Phi(w)}$
(9) Return $\{\hat{\pi}_e\}$

# 4 A CLOSER LOOK AT OUR EXAMPLE

Consider the CEG in Figure 2 and let the 16 edges be labelled $e_i$ in the same order as the $\{\theta_i\}$ thereon. In Examples 1 to 3 we showed how to create and use a Transporter CEG without concerning ourselves with a context. We now add that context and suppose that this CEG represents a Treatment regime for a serious medical condition, and the edges carry the meanings given in Table 2:

Table 2: Edge descriptors

| Edge | Description |
|---|---|
| $e_1$ | Not critical – Treatment prescribed I |
| $e_2$ | Liver failure – Treatment ... II |
| $e_3$ | Liver & Kidney failure – Treatment ... II |
| $e_4$ | Responds to I – Full recovery |
| $e_5$ | No response to I – Surgery prescribed III |
| $e_6, e_8$ | Responds to II – Surgery ... III |
| $e_7, e_9$ | No response to II – Surgery ... IV |
| $e_{10}$ | Recovery – Lifetime monitoring |
| $e_{11}$ | Recovery – Lifetime medication |
| $e_{12}, e_{13}$ | Death in surgery |
| $e_{14}$ | Survives surgery IV – Treatment ... V |
| $e_{15}$ | Recovery – Lifetime on treatment V |
| $e_{16}$ | No response to V – Dies |

As alluded to in section 2, it is not possible to represent this regime efficiently as a BN, nor yet as a context-specific BN, given that the asymmetry of the problem does not just lie in it having asymmetric sample space structures. By equating the descriptions of edges $e_4$ and $e_{10}$; edges $e_{11}$ and $e_{15}$; and edges $e_{12}, e_{13}$ and $e_{16}$, we can however **approximate** the problem with a 4-variable BN; where $X_1$ *Diagnosis and initial treatment* can take values corresponding to the outcomes $\{$*Not critical, Liver failure, Liver & Kidney failure*$\}$; $X_2$ *2nd treatment* to $\{$*None,* III, IV$\}$; $X_3$ *3rd treatment* to $\{$*None,* V$\}$; and $X_4$ *Response* to $\{$*Death, Partial recovery, Full recovery*$\}$. The BN for this approximation to the model is given in Figure 6.

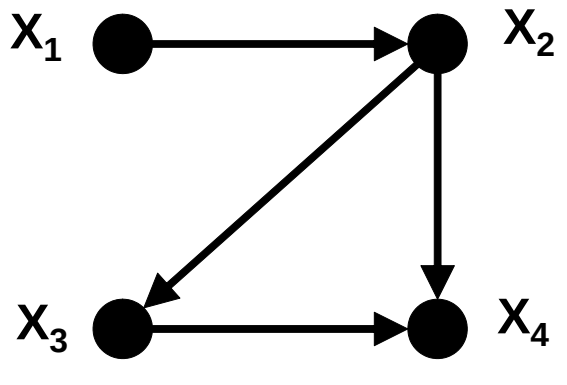

Figure 6: BN for our example

To store the model using a CEG requires 16 cells (corresponding to the 16 edges), but in this BN 27 cells (9 for the clique $\{X_1, X_2\}$ and 18 for $\{X_2, X_3, X_4\}$), 14 of which are storing the value zero.

The event $\Lambda$ in our example corresponds to the observation that a patient was not diagnosed with Liver **and** Kidney failure, and is still alive. Propagation of this event enables a practitioner to establish probability distributions for the possible histories of our patient. Note that it is only the fact that we can describe $\Lambda$ in such a simple manner that has allowed us to approximate the problem with the BN in Figure 6.

Propagating of the event $\Lambda$ using a simple Junction Tree algorithm on the cliques of the BN takes a minimum of 43 operations. Propagation on the CEG using our algorithm requires 32 operations (corresponding to 16 *backward* edges, 6 *backward* vertices and 10 *forward* edges). So even in this simple example, using the CEG is more efficient than the BN. The efficiency here is due mainly to the fact that the clique probability tables contain many zeros. This is reflected in the CEG by the $w_0 \to w_\infty$ paths not all having the same length. It is this form of asymmetry in a model that context-specific BNs do not cope with adequately, and why CEGs are a better structure for use with this type of problem.

The problems in which the algorithm described above are most efficient are when the CEG structure is known to be *simple*. To store the probability tables for the CEG requires only $N = \#(W(C)) + \#(E(C)) < 2\#(E(C))$ cells. In this case the *collect* step involves only $N$ calculations and the topology of the CEG is valid so that in particular the original probability table structure can be preserved. The potential product necessitates only a single distribute step which again only involves at most $N$ calculations. For large trees with much of the type of subtree symmetry discussed above the propagation is extremely fast.

It is worth quickly looking at a very simple example arising from model selection in graphical or partition model problems, an area currently attracting some interest: Consider a model with random variables $X_1, \ldots X_n$, where $X_1$ with $M = {}^1/_2(n-1)(n-2)$ possible states, determines which pair of binary variables from $\{X_2, \ldots X_n\}$ are dependent, all other variables from $\{X_2, \ldots X_n\}$ being independent of each other and of the pair determined. The CEG of this model has at most $M(1+2n)$ edges and $2 + Mn$ positions, whereas the BN is a single clique requiring $M \times 2^{n-1}$ cells for storage. As the number of operations required for propagation on both the BN and the CEG is of the same order of magnitude as the number of cells required for storage, it is clear that the CEG is far more efficient in this example.

## 5 DISCUSSION

There are several advantages of this method over the coding of this type of problem through a BN. Firstly, the calculated probabilities can be projected back on to the edges of the elicited tree, so that the consequences of inferences given different types of information can be immediately appreciated by the practi-tioner. Secondly, the accommmodation of data in the form of a compatible observation is much more general than the accommodation of subsets of observations from a predetermined set of random variables, so the CEG provides a more flexible framework for propagation, particularly when data is contingently censored. Thirdly, there are efficiency gains as outlined above. We intend to show how great these gains can be for very large problems in a later paper.

Note also that, as is the case with the triangulation step in BN-based algorithms, there are faster algorithms (Thwaites 2008) than the one described above, although they lose some of this algorithm's generality. Our algorithms are currently being coded by Cowell within freely available software, and will be available shortly.

Of course BNs provide a simpler representation of more symmetric problems and should always be preferred when the three contingencies are not satisified. The CEG does not provide a universal improvement over the BN for propagation. In particular in problems when the underlying BN is decomposable but the CEG is not simple the BN propagation can be much more efficient. But in highly asymmetric problems, the CEG should definitely be a first choice.

It should be noted that it is also possible to define a *dynamic* analogue of the CEG, and our investigation of these suggests that a *time-sliced* CEG (analogous to a *time-sliced* BN) will be an ideal vehicle for a dynamic updating algorithm. We hope to report on these developments in the near-future.

## APPENDIX

We claim that:

$$\begin{aligned}\hat{\pi}_e(w' \mid w) &\triangleq \pi(\Lambda(e(w,w')) \mid \Lambda, \Lambda(w)) \\ &= \begin{cases} \frac{\tau_e(w' \mid w)}{\Phi(w)} & \text{if } e(w,w') \in E_\Lambda \\ 0 & \text{if } e(w,w') \notin E_\Lambda \end{cases}\end{aligned}$$

**Proof:**

For a CEG $C$, and $C-$compatible event $\Lambda$, let $T$ be the tree associated with $C$, $T_\Lambda$ the tree associated with $C_\Lambda$, and $T_{(\Lambda)}$ the subtree of $T$ containing only those root-to-leaf paths in $\Lambda$. $T_{(\Lambda)}$ differs from $T_\Lambda$ in that the former retains the edge-probabilities from $T$.

Consider a position $w \in C$ ($w \in C_\Lambda$) corresponding to a set of vertices $\{v_i\} \in T$. Then the subtrees rooted in each $v_i$ are identical both in topology and in their edge-probabilities.

If there is a subpath $\mu(w_0, w)$ which is **not** part of a $w_0 \to w_\infty$ path in $\Lambda$ (ie. $\mu(w_0, w)$ exists in $C$, but **not** in $C_\Lambda$) then there will exist a subset of $\{v_i\}$ which does not exist in $T_\Lambda$ (or $T_{(\Lambda)}$). We split the set $\{v_i\}$ into:

$\{v_i\}_{i\in I}$ vertices existing in $T_\Lambda$
$\{v_i\}_{i\in J}$ vertices **not** existing in $T_\Lambda$

Because $\Lambda$ is $C-$compatible, the subtrees in $T_{(\Lambda)}$ rooted in each $v_i \in \{v_i\}_{i\in I}$ are also identical both in topology and in their edge-probabilities that they retain from $T$.

Suppose there exists an edge $e(w,w')$ in $C$, then for each $v_i \in \{v_i\}$, there exists an edge $e(v_i, v_i')$ in $T$ corresponding to this edge. Note that:

$$\Lambda(w) = \bigcup_{i\in I\cup J} \Lambda(v_i)$$
$$\Lambda(e(w,w')) = \bigcup_{i\in I\cup J} \Lambda(e(v_i, v_i'))$$
$$\pi_e(v_i' \mid v_i) = \pi_e(w' \mid w) \quad \forall i \in I \cup J$$

and since the subtrees in $T_{(\Lambda)}$ rooted in each $v_i \in \{v_i\}_{i\in I}$ are identical, we also have:

$$\pi(\Lambda \mid \Lambda(v_i)) = \pi(\Lambda \mid \Lambda(v_j))$$
$$\pi(\Lambda, \Lambda(e(v_i, v_i')) \mid \Lambda(v_i)) = \pi(\Lambda, \Lambda(e(v_j, v_j')) \mid \Lambda(v_j))$$
$$\text{for } i,j \in I$$

[$\pi(\Lambda \mid \Lambda(v_i))$ is the sum of the probabilities of all the $\mu(v_i, v_{leaf})$ subpaths in $T_{(\Lambda)}$, and $\pi(\Lambda, \Lambda(e(v_i, v_i')) \mid \Lambda(v_i))$ is the sum of the probabilities of all the $\mu(v_i, e(v_i, v_i'), v_i', v_{leaf})$ subpaths in $T_{(\Lambda)}$]

So:

$$\begin{aligned}\hat{\pi}_e(w' \mid w) &= \pi(\Lambda(e(w,w')) \mid \Lambda, \Lambda(w)) \\ &= \frac{\pi(\Lambda, \Lambda(w), \Lambda(e(w,w')))}{\pi(\Lambda, \Lambda(w))} \\ &= \frac{\pi(\Lambda, \bigcup_{i\in I\cup J}[\Lambda(v_i), \Lambda(e(v_i, v_i'))])}{\pi(\Lambda, \bigcup_{i\in I\cup J} \Lambda(v_i))}\end{aligned}$$

(an expression evaluated on $T$)

since $\Lambda(v_i) \cap \Lambda(e(v_j, v_j')) = \phi$ for $i \neq j$

$$= \frac{\sum_{i\in I\cup J} \pi(\Lambda, \Lambda(v_i), \Lambda(e(v_i, v_i')))}{\sum_{i\in I\cup J} \pi(\Lambda, \Lambda(v_i))}$$

But $\Lambda \cap \Lambda(v_i) = \phi$ for $v_i \in \{v_i\}_{i\in J}$, so this equals:

$$\begin{aligned}&\frac{\sum_{i\in I} \pi(\Lambda, \Lambda(v_i), \Lambda(e(v_i, v_i')))}{\sum_{i\in I} \pi(\Lambda, \Lambda(v_i))} \\ &= \frac{\sum_{i\in I} \pi(\Lambda, \Lambda(e(v_i, v_i')) \mid \Lambda(v_i))\ \pi(\Lambda(v_i))}{\sum_{i\in I} \pi(\Lambda \mid \Lambda(v_i))\ \pi(\Lambda(v_i))} \\ &= \frac{\pi(\Lambda, \Lambda(e(v_j, v_j')) \mid \Lambda(v_j))\ \sum_{i\in I} \pi(\Lambda(v_i))}{\pi(\Lambda \mid \Lambda(v_j))\ \sum_{i\in I} \pi(\Lambda(v_i))}\end{aligned}$$

for any $v_j \in \{v_i\}_{i\in I}$

$$= \frac{\pi(\Lambda, \Lambda(e(v_j, v_j')) \mid \Lambda(v_j))}{\pi(\Lambda \mid \Lambda(v_j))}$$

for any $v_j \in \{v_i\}_{i\in I}$

Turning our attention to the terms in the algorithm, we claim that $\Phi(w) = \pi(\Lambda \mid \Lambda(v_i))$ and $\tau_e(w' \mid w) = \pi(\Lambda, \Lambda(e(v_i, v_i')) \mid \Lambda(v_i))$ ($v_i \in \{v_i\}_{i\in I}$) for all $w$, $e(w,w') \in C_\Lambda$, where $\{v_i\}_{i\in I}$ is the set of vertices in $T_{(\Lambda)}$ corresponding to $w$. We prove this by induction:

**Step 1.**

Consider positions $w \in W(-1)$. Then:

$$\Phi(w) = \sum_e \tau_e(w_\infty \mid w) = \sum_e \pi_e(w_\infty \mid w)$$
$$= \sum_e \pi_e(v_{leaf} \mid v_i) \qquad \text{in } T_{(\Lambda)}$$

for any $v_i \in \{v_i\}_{i \in I}$

$$= \pi(\Lambda \mid \Lambda(v_i))$$

**Step 2.**

Suppose $w$ is such that all of its outgoing edges terminate in positions $\{w'\}$ for which $\Phi(w') = \pi(\Lambda \mid \Lambda(v_i'))$. Then:

$$\Phi(w) = \sum_e \tau_e(w' \mid w) = \sum_e \pi_e(w' \mid w)\ \Phi(w')$$
$$= \sum_e \pi_e(v_i' \mid v_i)\ \pi(\Lambda \mid \Lambda(v_i'))$$

for any $v_i \in \{v_i\}_{i \in I}$

$$= \sum_e \pi(\Lambda(e(v_i, v_i')) \mid \Lambda(v_i))\ \pi(\Lambda \mid \Lambda(v_i'))$$

But $\Lambda(v_i') = \Lambda(e(v_i, v_i')) \subset \Lambda(v_i)$ in a tree, so this equals:

$$\sum_e \pi(\Lambda(e(v_i, v_i')), \Lambda(v_i') \mid \Lambda(v_i))$$
$$\times\ \pi(\Lambda \mid \Lambda(v_i), \Lambda(e(v_i, v_i')), \Lambda(v_i'))$$
$$= \sum_e \pi(\Lambda, \Lambda(e(v_i, v_i')), \Lambda(v_i') \mid \Lambda(v_i))$$
$$= \sum_e \pi(\Lambda, \Lambda(e(v_i, v_i')) \mid \Lambda(v_i))$$
$$= \pi(\Lambda, \Lambda(v_i) \mid \Lambda(v_i)) = \pi(\Lambda \mid \Lambda(v_i))$$

Hence:

$$\tau_e(w' \mid w) = \pi_e(w' \mid w)\ \Phi(w')$$
$$= \pi_e(v_i' \mid v_i)\ \pi(\Lambda \mid \Lambda(v_i'))$$

for any $v_i \in \{v_i\}_{i \in I}$

$$= \pi(\Lambda(e(v_i, v_i')) \mid \Lambda(v_i))\ \pi(\Lambda \mid \Lambda(v_i'))$$
$$= \ldots = \pi(\Lambda, \Lambda(e(v_i, v_i')) \mid \Lambda(v_i))$$

We now combine our two results to give:

$$\hat{\pi}_e(w' \mid w) = \frac{\pi(\Lambda, \Lambda(e(v_j, v_j')) \mid \Lambda(v_j))}{\pi(\Lambda \mid \Lambda(v_j))}$$
$$= \frac{\tau_e(w' \mid w)}{\Phi(w)} \qquad \square$$

**Acknowledgements**

This work has been partly funded by the EPSRC as part of the project *Chain Event Graphs: Semantics and Inference.*